\documentclass[conference]{IEEEtran}
\IEEEoverridecommandlockouts

\usepackage{cite}
\usepackage{graphicx}
\usepackage{amsmath}
\usepackage{amssymb}
\usepackage{booktabs}
\usepackage[hyphens]{url}
\usepackage{flushend}
\newcommand{\mN}{2{,}537}
\newcommand{\mFmWin}{50.3}
\newcommand{\mPureFM}{1.206}
\newcommand{\mPureSpec}{1.233}
\newcommand{\mFmSpec}{0.027}
\newcommand{\mFmSpecLo}{0.007}
\newcommand{\mFmSpecHi}{0.048}
\newcommand{\mRand}{1.225}
\newcommand{\mOrc}{1.072}
\newcommand{\mOrcBestA}{0.503}
\newcommand{\mOrcBest}{1.059}
\newcommand{\mPrec}{61.0}
\newcommand{\mRest}{45.3}
\newcommand{\mGap}{30}
\newcommand{\mKcalOne}{98}
\newcommand{\mLrDiff}{+0.002}
\newcommand{\mLrDiffSd}{0.004}
\newcommand{\mLrBetter}{23}
\newcommand{\mOneShare}{31.6}
\newcommand{\mOneSaved}{68.4}
\newcommand{\mOneMase}{1.179}
\newcommand{\mOneD}{-0.027}

\newcommand{\mOneLo}{-0.041}
\newcommand{\mOneHi}{-0.013}

\newcommand{\mOneRD}{-0.027}

\newcommand{\mOneRSd}{0.008}
\newcommand{\mOneFrac}{100}
\newcommand{\mOneRShare}{31.4}
\newcommand{\mOneRShSd}{9.4}

\newcommand{\mTwoD}{-0.011}

\newcommand{\mTwoLo}{-0.026}
\newcommand{\mTwoHi}{+0.004}

\newcommand{\mTwoRD}{-0.004}

\newcommand{\mTwoFrac}{63}

\newcommand{\mTwoRPlo}{5}
\newcommand{\mTwoRPhi}{44}

\newcommand{\mSEFrac}{59}

\newcommand{\mCVFrac}{50}

\newcommand{\mSACShare}{19.1}

\newcommand{\mSACD}{-0.023}

\newcommand{\mSACLo}{-0.042}
\newcommand{\mSACHi}{-0.005}

\newcommand{\mSACFrac}{97}

\newcommand{\mLRShare}{51.6}

\newcommand{\mLRD}{-0.034}

\newcommand{\mLRLo}{-0.046}
\newcommand{\mLRHi}{-0.021}

\newcommand{\mLRRD}{-0.029}

\newcommand{\mLRmD}{-0.027}

\newcommand{\mLRmLo}{-0.040}
\newcommand{\mLRmHi}{-0.013}

\newcommand{\pN}{3{,}055}
\newcommand{\pFmWin}{59.5}

\newcommand{\pFmSpec}{0.118}

\newcommand{\pLrDiff}{-0.010}
\newcommand{\pLrDiffSd}{0.006}
\newcommand{\pLrBetter}{93}

\newcommand{\pOneD}{-0.005}

\newcommand{\pOneLo}{-0.015}
\newcommand{\pOneHi}{+0.005}

\newcommand{\pOneRD}{-0.014}

\newcommand{\pOneFrac}{99}

\newcommand{\pLRShare}{52.0}

\newcommand{\pLRD}{-0.026}

\newcommand{\pLRLo}{-0.036}
\newcommand{\pLRHi}{-0.016}

\newcommand{\mSEthr}{0.398}
\newcommand{\mCVthr}{0.407}
\newcommand{\mSACthr}{0.994}
\newcommand{\pTrafficKOneSh}{100.0}

\newcommand{\pMFourKOneSh}{40.2}

\newcommand{\costEps}{7}
\newcommand{\costBE}{11}
\newcommand{\mOneRShareShort}{31}

\begin{document}

\title{The Complexity Router: Cost-Aware Foundation Model Deployment for Time Series Forecasting\thanks{\copyright~2026 IEEE. Personal use of this material is permitted. Permission from IEEE must be obtained for all other uses, in any current or future media, including reprinting/republishing this material for advertising or promotional purposes, creating new collective works, for resale or redistribution to servers or lists, or reuse of any copyrighted component of this work in other works.}}

\author{
\IEEEauthorblockN{Kavin Soni}
\IEEEauthorblockA{Google\\
Mountain View, CA, USA\\
kavinsoni@google.com}
\and
\IEEEauthorblockN{Vamshi Guduguntla}
\IEEEauthorblockA{Google\\
Mountain View, CA, USA\\
vamshigud@google.com}
\and
\IEEEauthorblockN{Debanshu Das}
\IEEEauthorblockA{Google\\
Mountain View, CA, USA\\
debanshu@google.com}
}

\maketitle

\begin{abstract}
Foundation models (FMs) for time series forecasting offer zero-shot
generalization, but their inference cost can be significantly higher
than that of lightweight supervised specialists, and practitioners lack
guidance on when that cost is justified. We introduce the Complexity
Router, a pre-inference router that assigns each series to an FM or to a
specialist using four features of its history: spectral entropy,
coefficient of variation, seasonal autocorrelation and trend strength.
No model is run at routing time. The FM and the specialist are chosen on
a calibration split before deployment, and we report results on held-out
series over 100 repeated calibration and holdout splits, with each model
trained once. On M4 Daily, where the FM and a linear
specialist are close in accuracy, a 1-of-4 vote rule improves on
always-FM (MASE \mOneMase{} against \mPureFM{}) while sending only
\mOneShare\% of series to the FM, and does so in all 100 repeated routing splits. A logistic regression on the same features matches it at equal FM share
on M4 and beats it at equal share in \pLrBetter\% of splits on a mixed
Traffic and M4 portfolio. With the router's own overhead counted, savings
appear once the FM costs more than roughly \costBE{} times a specialist call.
An oracle ranking bounds the remaining headroom.
\end{abstract}


\section{Introduction}

Foundation models (FMs) for time series, large pre-trained models
capable of zero-shot inference, offer an alternative to task-specific
supervised pipelines by removing per-series training and feature
engineering. This makes them attractive for cold-start series with too
little history to train a specialist and for long-tail catalogs where
maintaining per-series pipelines is impractical. Once history is
available, however, zero-shot generalization comes at an operational
cost. FMs require GPU infrastructure and incur far higher inference
latency than supervised specialists such as lightweight linear and
neural models. For high-throughput industrial systems forecasting
millions of series, universal FM deployment can be computationally
infeasible, and on some datasets it is not even the most accurate choice.

We address this by deciding, for each series, whether to pay for the FM
before any model is invoked. The Complexity Router uses four features
computed only from the training history: spectral entropy, coefficient
of variation, seasonal autocorrelation and trend strength. We combine
them using either an auditable vote rule with thresholds read off
calibration deciles or a logistic regression. Computing the features
takes about a millisecond per series.

We evaluate under a protocol that mirrors deployment. The FM and the
specialist are chosen on a calibration split, thresholds and classifiers
are fit on the same split, and every result is reported on held-out
series, over 100 repeated random splits as well as one headline split.
We emphasize the setting in which routing is hardest, M4 Daily, where the
FM and the specialist are close in accuracy, and use a mixed
Traffic and M4 portfolio as a secondary test. Our contributions are:
(1)~an evaluation protocol for FM-versus-specialist routing that uses
only information available before deployment, with repeated splits
and a measurement of how much per-series hindsight selection overstates
accuracy (Section~\ref{sec:protocol});
(2)~evidence that, within M4 Daily, a 1-of-4 vote rule improves on
always-FM in all 100 repeated routing splits with about one third of the
FM calls, that a logistic regression on the same features matches it at
the same cost, and that the learned router beats the vote rule at equal
cost on a mixed Traffic and M4 portfolio
(Section~\ref{sec:router});
(3)~a cost analysis that includes the router's own overhead and gives
the cost ratio above which routing pays off; and
(4)~ablations over the vote threshold and over single features, and an
oracle bound on the remaining headroom.

\section{Background and Related Work}

\paragraph{Time Series Foundation Models}
Recent large-scale pre-trained models for forecasting, including
TimesFM~\cite{das2023timesfm} and Chronos~\cite{ansari2024chronos},
demonstrate strong zero-shot generalization across domains. Systematic
benchmarks such as GIFT-Eval~\cite{aksu2024gifteval} and
FoundTS~\cite{li2024foundts} show that specialist models frequently
match or outperform FMs on structured or low-complexity series,
motivating guidance on when FM deployment is warranted.

\paragraph{Ensemble and Hybrid Methods}
Modi and Pan~\cite{modi2025ensemble} show that bagging, boosting and
statistical ensembles improve FM robustness, but all constituent models
must run at inference time. Hybrid architectures that combine
statistical and deep components~\cite{smyl2020esrnn} produce fixed
pipelines that cannot adapt the choice of model per series.

\paragraph{Feature-Based Model Selection}
Choosing a forecasting method from series features has a long history,
within the broader algorithm selection problem~\cite{rice1976algorithm}.
FFORMA~\cite{montero2020fforma} and
FFORMS~\cite{talagala2018fforms} learn meta-models that map series features
to combination weights or to the best method, and instance-space
analysis~\cite{kang2017instance} relates features to algorithm
performance. FFORMA averages the forecasts of every candidate, so it
runs the full pool at inference time, which is affordable for
statistical methods but not when a candidate is a GPU-bound FM. FFORMS,
like our router, selects a single method before any forecast is produced;
we differ in the candidate pool, a GPU-bound FM against a cheap
specialist, and in treating the choice as a cost-accuracy trade-off
evaluated with calibration-time information only. Wang et
al.~\cite{wang2025spectral} show that spectral predictability
stratifies model-family performance at the dataset level.

\paragraph{Routing and Cascades}
In language modeling, routers and cascades decide which model to query
to save cost: FrugalGPT~\cite{chen2023frugalgpt}, Hybrid
LLM~\cite{ding2024hybridllm} and RouteLLM~\cite{ong2024routellm} send
easy queries to a cheaper model. We apply the same idea to forecasting
with an FM and a cheap specialist. The methods we use, hand-set
feature thresholds and a logistic regression, are deliberately simple
and established. Our contribution is the setting, a pre-inference
decision from cheap history features, the evaluation protocol, and an
explicit accounting of when it pays off.

\section{Experimental Setup}

\subsection{Datasets}
We use four datasets (Table~\ref{tab:datasets}). M4
Daily~\cite{makridakis2018m4} contains 4{,}227 heterogeneous demand series
from diverse industries and is our primary routing benchmark: the FM and
the best specialist are close in accuracy, so which model to use is a
real question for most series. Traffic is drawn from the California
PeMS system~\cite{caltrans_pems}: 862 hourly sensor series with strong
diurnal and weekly periodicity, on which the FM dominates. Together with
M4 it forms a 5{,}089-series mixed portfolio. ETTh1 is an hourly
electricity transformer temperature dataset~\cite{zhou2021informer}
and Exchange Rate contains daily foreign exchange rates across eight
currency pairs~\cite{lai2018modeling}. With 7 and 8 series they are
too small for routing and serve only as a regime check
(Section~\ref{sec:protocol}).

\begin{table}[t]
\caption{Datasets. $S$ is the seasonal period used by the routing
features (the MASE denominator uses 24 for Traffic and ETTh1), $L$ the
lookback window and $H$ the forecast horizon.}
\label{tab:datasets}
\centering
\footnotesize
\setlength{\tabcolsep}{4pt}
\begin{tabular}{lrlrrrl}
\toprule
\textbf{Dataset} & \textbf{Series} & \textbf{Freq.}
  & $S$ & $L$ & $H$ & \textbf{Regime} \\
\midrule
M4 Daily       & 4{,}227 & Daily  & 7   & 30  & 14  & Demand     \\
Traffic (PeMS) & 862     & Hourly & 168 & 168 & 168 & Transport. \\
ETTh1          & 7       & Hourly & 24  & 96  & 24  & Energy     \\
Exchange Rate  & 8       & Daily  & 7   & 96  & 96  & Finance    \\
\bottomrule
\end{tabular}
\end{table}

\subsection{Models}
Two foundation models run in zero-shot mode with no per-series training
and receive only the raw context window of length $L$: TimesFM 2.5
(200M)~\cite{das2023timesfm}, with maximum context set to $L$ and input
normalization enabled, and Chronos T5-Base (200M)~\cite{ansari2024chronos},
using the median of sampled trajectories as the point forecast. Two
supervised specialists are trained independently on each dataset:
PatchTST~\cite{nie2023patchtst} and DLinear~\cite{zeng2023ltsflinear}.
Table~\ref{tab:hparams} lists their hyperparameters. Both are trained
with Adam (learning rate $10^{-3}$) on MSE loss, hold out the last 20\%
of each series' training history for validation, and stop early on validation loss.
Inputs are $\log(1+x)$ transformed for Traffic and z-score normalized
for M4; predictions are inverted to the original scale before
evaluation. Each specialist is trained once per dataset on the training-period windows of
all its series, including the series later held out for routing
evaluation; test periods are never used. The calibration and holdout split
therefore concerns the routing step only, as in deployment, where
specialists are fit on history and routing is applied to the series being
forecast. The routing splits use seed 42, but neural training and Chronos
sampling are not seeded, so independent runs of the same model differ
(Section~\ref{sec:protocol}).

\begin{table}[t]
\caption{Specialist hyperparameters. Values separated by a slash
apply to Traffic / M4 where they differ.}
\label{tab:hparams}
\centering
\footnotesize
\begin{tabular}{@{}lp{0.72\columnwidth}@{}}
\toprule
\textbf{Model} & \textbf{Configuration} \\
\midrule
PatchTST & $d_{\text{model}}=128$, 16 heads, 3 encoder layers,
           $d_{\text{ff}}=256$, dropout 0.2; patch length 16,
           stride 8 (4 and 2 on M4, giving 14 patches for $L=30$);
           batch 32 / 256; up to 30 / 5 epochs, patience 5 / 2 \\
\addlinespace
DLinear  & moving-average kernel 25 for trend-seasonal
           decomposition; batch 32 / 256; up to 20 epochs,
           patience 3 \\
\bottomrule
\end{tabular}
\end{table}

\subsection{Evaluation Metric}
All models are evaluated with the Mean Absolute Scaled Error
(MASE)~\cite{hyndman2006mase}, which normalizes forecast error by the
in-sample seasonal naive error, using a period of 24 for Traffic and ETTh1 and 7
for M4 and Exchange, computed over the full training history. Each series
has one forecast origin: the first $H$ steps after the training period.
Lower is better and 1.0 equals the seasonal naive baseline. Because we
scale M4 Daily by the lag-7 naive error, our MASE values are not
comparable with M4 competition figures, which use lag 1. All reported MASE values are averages over
series.

\section{Evaluation Protocol and Model Accuracy}
\label{sec:protocol}

Table~\ref{tab:results} reports mean MASE for the four models over all
series. On Traffic the FMs are far more accurate than the specialists
(TimesFM 2.5 at 0.621 against DLinear at 1.118). On M4 Daily the gap
nearly vanishes: Chronos (1.214), TimesFM 2.5 (1.234) and DLinear (1.247)
lie within 0.033 of each other, and PatchTST is clearly worse (1.771).
Whether the FM is worth its cost is therefore a dataset-dependent
question, and on M4 it is a close call for most series.

\begin{table}[t]
\caption{Mean MASE over all series. Bold marks the best model per
dataset. Lower is better.}
\label{tab:results}
\centering
\small
\begin{tabular}{llcc}
\toprule
\textbf{Model} & \textbf{Class} & \textbf{Traffic} & \textbf{M4 Daily} \\
\midrule
TimesFM 2.5 & FM & \textbf{0.621} & 1.234 \\
Chronos & FM & 0.721 & \textbf{1.214} \\
PatchTST & Spec. & 1.307 & 1.771 \\
DLinear & Spec. & 1.118 & 1.247 \\
\bottomrule
\end{tabular}

\end{table}

\paragraph{Splits and model choice}
We split the series of each dataset into a calibration set (40\%) and a
holdout set (60\%). For M4 alone this gives 1{,}690 calibration and
2{,}537 holdout series, and for the mixed portfolio 2{,}034 and 3{,}055,
stratified by dataset. On the calibration set only, we pick for each
dataset the FM and the specialist with the lowest mean MASE; on the
headline split (seed 42) this selects Chronos and DLinear for M4 and
TimesFM 2.5 and DLinear for Traffic. The MASE of the FM class for a
series is that of its selected FM, and likewise for the specialist
class. Thresholds and classifiers are fit on the calibration set. To
check that conclusions do not hinge on one split, we repeat the whole
procedure, including model choice, thresholds and classifier fitting,
on 100 further random splits and report the mean and spread.

\paragraph{Why not select per series}
Taking, for each series, the better of two FMs by realized test error
cannot be done in deployment, because the realized error is unknown until
the horizon has passed, and running both models does not reveal it in
advance. It also overstates accuracy. On the M4 holdout it lowers the
FM-class MASE from 1.206 to 1.065. We therefore evaluate every method
with calibration-time model choices.

\subsection{Regime Check on ETTh1 and Exchange}
ETTh1 and Exchange are too small for decile analysis or routing, so we
use them only to test whether FM dominance varies across regimes
(Table~\ref{tab:regime}). Neural training and Chronos sampling are not
seeded, so independent runs of the same model differ: DLinear on Traffic
scores 1.165 in an earlier run and 1.118 in the later one. The effect is
much larger on Exchange, where PatchTST scored 2.46, 2.82, 3.76 and 4.65 across
four runs. We therefore compare the regimes only qualitatively.

\begin{figure}[t]
  \centering
  \includegraphics[width=\columnwidth]{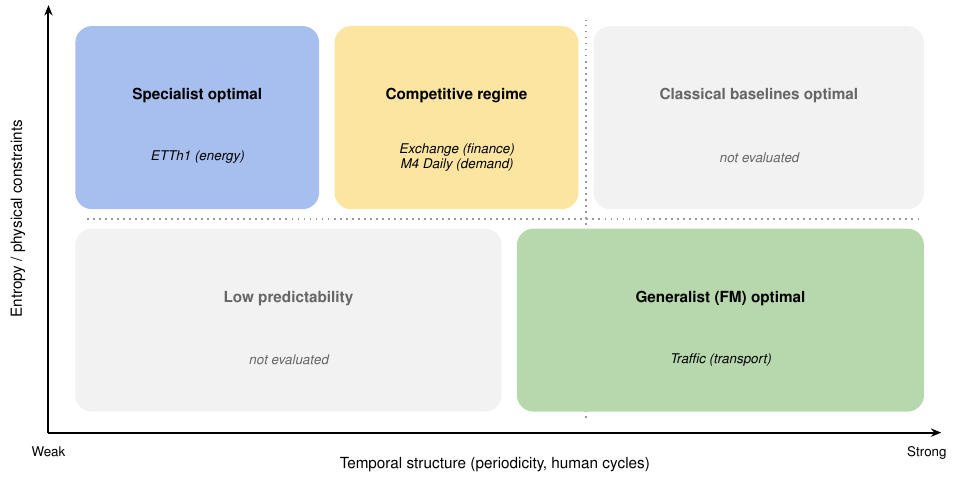}
  \caption{Regime map summarizing Tables~\ref{tab:results}
  and~\ref{tab:regime}. Optimal marks the more accurate model class on
  the evaluated datasets; dataset positions along the axes are
  qualitative. ETTh1 and Exchange rest on 7 and 8 series, and the two
  grey regions were not evaluated.}
  \label{fig:regimemap}
\end{figure}

On ETTh1 both specialists beat both FMs: PatchTST (0.856) and DLinear
(0.927) against TimesFM 2.5 (1.046) and Chronos (1.102). On Exchange, a
low-signal, near-random-walk regime, TimesFM 2.5 has the lowest score
(2.167) and Chronos scores 2.569, but the specialists' scores vary so much
across runs (DLinear 2.44 to 2.55, PatchTST 2.46 to 4.65) that with 8
series we cannot rank the two classes. Together with
Table~\ref{tab:results}, FM dominance is strong on Traffic, marginal on M4
Daily, absent on ETTh1 and unsettled on Exchange.
Fig.~\ref{fig:regimemap} summarizes these results on a regime map. The
winner in each region is measured; the axis positions are qualitative.

\begin{table}[t]
\caption{Regime check: dataset-level MASE. Values are single runs, except
the Exchange specialists, which show the range over independent runs. Bold
marks the lowest single value per dataset. Lower is better.}
\label{tab:regime}
\centering
\small
\begin{tabular}{llcc}
\toprule
\textbf{Model} & \textbf{Class} & \textbf{ETTh1} & \textbf{Exchange} \\
\midrule
TimesFM 2.5 & FM    & 1.046 & \textbf{2.167} \\
Chronos     & FM    & 1.102 & 2.569 \\
PatchTST    & Spec. & \textbf{0.856} & 2.46 to 4.65 \\
DLinear     & Spec. & 0.927 & 2.44 to 2.55 \\
\bottomrule
\end{tabular}
\end{table}

\subsection{Inference Cost}
Table~\ref{tab:throughput} documents throughput. FMs run on GPU with P95
latencies of 283\,ms and 1{,}764\,ms; DLinear and PatchTST run on CPU at
0.13\,ms and 9.03\,ms. Measured throughput differs by more than
$2{,}000\times$ between TimesFM 2.5 and DLinear, but the two run on
different hardware and FM throughput may reflect unbatched inference, so we
treat the cost ratio $C_{\text{FM}}/C_{\text{spec}}$ as a parameter and report
results across a range (Section~\ref{sec:cost}). Extracting the four routing features on a single CPU core takes about
0.9\,ms on M4 and 2.7\,ms on Traffic (14{,}035 points), so we include
this overhead in the cost model.

\begin{table}[t]
\caption{Inference throughput. FMs run on GPU, specialists on CPU.}
\label{tab:throughput}
\centering
\small
\begin{tabular}{llrr}
\toprule
\textbf{Model} & \textbf{HW} & \textbf{P95 Lat.}
  & \textbf{Series/s} \\
\midrule
TimesFM 2.5 (200M) & T4   & 283\,ms     & 3.5     \\
Chronos (200M)     & T4   & 1{,}764\,ms & 0.6     \\
PatchTST           & Xeon & 9.03\,ms    & 111     \\
DLinear            & Xeon & 0.13\,ms    & 7{,}870 \\
\bottomrule
\end{tabular}
\end{table}

\section{The Complexity Router}
\label{sec:router}

The router decides, for each series, whether the expected accuracy gain
from the FM justifies its cost. It never consults model outputs at
routing time.

\subsection{Routing Features}
For a training history $y_1,\dots,y_n$ with mean $\mu$, standard
deviation $\sigma$ and seasonal period $S$, the router computes
\textbf{spectral entropy}, the Shannon entropy of the normalized power
spectrum,
\begin{equation}
  H_{\text{spec}} = -\sum_k p_k \log p_k, \qquad
  p_k = \frac{|Y_k|^2}{\sum_j |Y_j|^2},
\end{equation}
where $Y_k$ is the real FFT of the series; the \textbf{coefficient of
variation}, $\mathrm{CV} = \sigma / \max(|\mu|, 10^{-5})$;
\textbf{seasonal autocorrelation} $\rho_S$, the Pearson correlation
between $y_t$ and $y_{t-S}$; and \textbf{trend strength}
$R^2_{\text{trend}}$, the coefficient of determination of an ordinary
least squares fit of $y_t$ on $t$. The features are computed on the full training history, which is longer
than the FM's context window: at least 93 points on M4 against $L=30$, and
14{,}035 on Traffic against $L=168$. Extraction takes about a millisecond per
series on one CPU core.

\begin{figure}[t]
  \centering
  \includegraphics[width=\columnwidth]{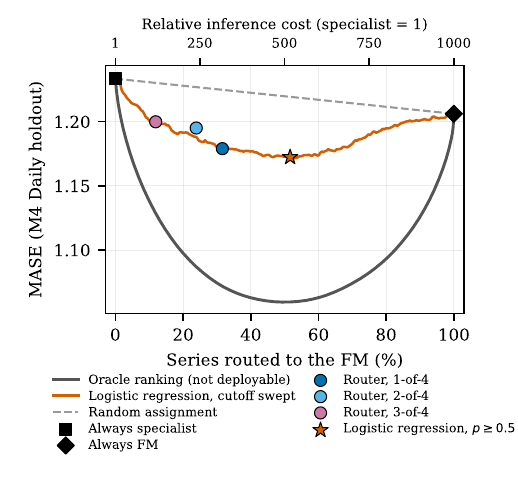}
  \caption{M4 Daily holdout MASE against the share of series routed to
  the FM (headline split). The oracle ranks series by true FM advantage
  and cannot be deployed. The orange curve sweeps the logistic
  regression's probability cutoff. The top axis assumes an FM-to-specialist
  cost ratio of 1000 and excludes router overhead.}
  \label{fig:pareto}
\end{figure}

\subsection{Two Ways to Combine the Features}
\begin{figure*}[t]
  \centering
  \includegraphics[width=\textwidth]{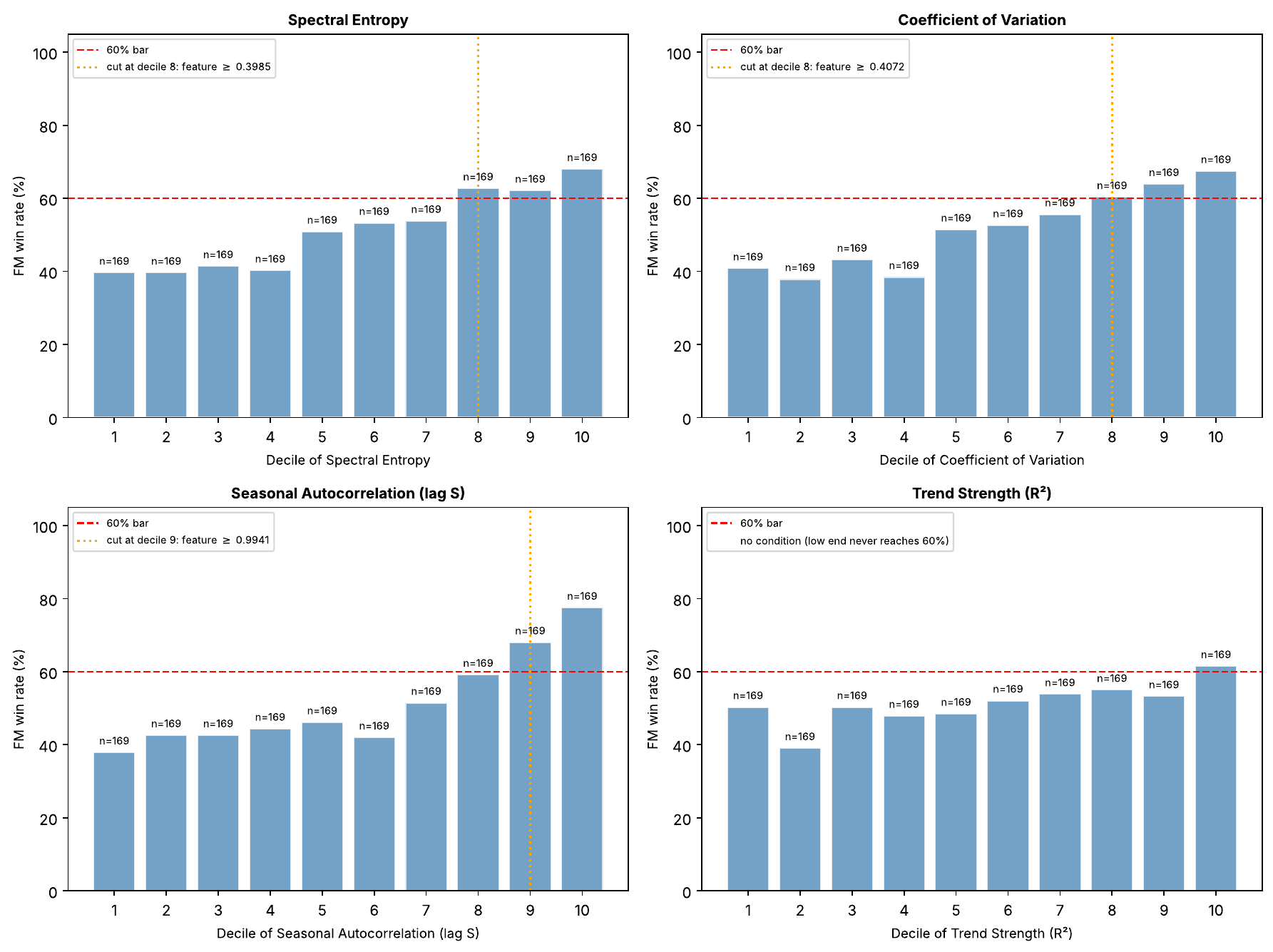}
  \caption{FM win rate by decile for the four routing features on the M4
  Daily calibration set ($n=1{,}690$, headline split). The red dashed
  line marks the 60\% bar and the dotted lines mark the derived cut
  points. The trend-strength condition targets low values, where no decile
  reaches the bar. Its top decile reaches 61.5\%, but that end is not
  used, so trend strength yields no condition on M4.}
  \label{fig:deciles}
\end{figure*}

\paragraph{Vote rule}
Thresholds are derived on the calibration set, where a series is labeled
FM-favorable if the selected FM achieves lower MASE than the selected
specialist. Each feature gets a
condition at the decile where the FM win rate first exceeds 60\%
(Fig.~\ref{fig:deciles}). A series is routed to the FM when at least $k$
of the four conditions hold, and $k$ is a cost dial. On M4 the headline
split gives spectral entropy $\geq \mSEthr$, CV $\geq \mCVthr$ and
seasonal autocorrelation $\geq \mSACthr$. The trend-strength condition is
defined for low trend strength, and none of the lowest deciles reaches
the 60\% bar. Only the top decile does (61.5\%), an end the rule does
not use, so the condition never fires on M4 and a 4-of-4 vote routes
nothing to the FM. We select $k$ on the calibration set as the
value with the lowest calibration MASE; this picks $k=1$ in \mKcalOne{} of
100 splits on M4.

\paragraph{Logistic regression}
We also train a logistic regression on the four standardized features
and the same calibration labels. It routes a series to the FM when its
predicted FM-win probability is at least 0.5. To compare it with the
vote rule at equal cost, we also report a version that sends to the
FM the same share of holdout series as the 1-of-4 rule, choosing them by
highest predicted probability; this uses no holdout outcomes.

\subsection{Results on M4 Daily}
Table~\ref{tab:m4} reports the M4 Daily holdout ($n=\mN$). The FM is
only slightly more accurate than DLinear (MASE \mPureFM{} against
\mPureSpec{}, difference \mFmSpec{}, 95\% CI $[\mFmSpecLo, \mFmSpecHi]$,
paired bootstrap with 1{,}000 resamples), and it wins on only \mFmWin\% of
series.

\begin{table*}[t]
\caption{M4 Daily holdout, headline split ($n=\mN$) and 100 repeated
splits. FM share is the share of series routed to the FM, with its
standard deviation over splits in parentheses. The first $\Delta$ column
is the paired difference in MASE from always-FM with a 95\% bootstrap
confidence interval; the second gives its mean and standard deviation
over splits; negative is better than always-FM. ``Better'' is the
percentage of splits in which the method beats always-FM. A 4-of-4 vote
and trend strength alone never route to the FM.}
\label{tab:m4}
\centering
\footnotesize
\setlength{\tabcolsep}{5pt}
\begin{tabular}{lrrlrr}
\toprule
\textbf{Method} & \textbf{FM share (\%)} & \textbf{MASE} & \textbf{$\Delta$ vs.\ always-FM [95\% CI]} & \textbf{$\Delta$, 100 splits} & \textbf{Better (\%)} \\
\midrule
Always specialist (DLinear) & 0.0 (0.0) & 1.233 & $+0.027$ [$+0.007$, $+0.048$] & $+0.034$ (0.008) & 0 \\
Always FM & 100.0 & 1.206 & --- & --- & --- \\
\midrule
Router, 1-of-4 vote & 31.6 (9.4) & 1.179 & $-0.027$ [$-0.041$, $-0.013$] & $-0.027$ (0.008) & 100 \\
Router, 2-of-4 vote & 23.9 (9.9) & 1.195 & $-0.011$ [$-0.026$, $+0.004$] & $-0.004$ (0.016) & 63 \\
Router, 3-of-4 vote & 11.9 (5.0) & 1.200 & $-0.006$ [$-0.026$, $+0.012$] & $+0.009$ (0.015) & 25 \\
\midrule
Spectral entropy alone & 24.1 (12.2) & 1.195 & $-0.011$ [$-0.025$, $+0.004$] & $-0.003$ (0.015) & 59 \\
CV alone & 24.2 (12.2) & 1.196 & $-0.010$ [$-0.025$, $+0.005$] & $+0.002$ (0.017) & 50 \\
Seasonal autocorr. alone & 19.1 (5.5) & 1.183 & $-0.023$ [$-0.042$, $-0.005$] & $-0.022$ (0.009) & 97 \\
\midrule
Logistic regression ($p\geq0.5$) & 51.6 (6.3) & 1.172 & $-0.034$ [$-0.046$, $-0.021$] & $-0.029$ (0.004) & 100 \\
Logistic regression (same FM share as 1-of-4) & 31.6 (9.4) & 1.179 & $-0.027$ [$-0.040$, $-0.013$] & $-0.025$ (0.009) & 100 \\
\bottomrule
\end{tabular}

\end{table*}

\paragraph{The vote rule improves on always-FM}
With $k=1$ the router sends \mOneShare\% of series to the FM and reaches
MASE \mOneMase{}, against \mPureFM{} for always-FM, a difference of
$\mOneD$ (95\% CI $[\mOneLo, \mOneHi]$) with \mOneSaved\% fewer FM
calls. Over 100 splits the mean difference is $\mOneRD$ (standard
deviation \mOneRSd) and the rule beats always-FM in \mOneFrac\% of
splits, with an average FM share of \mOneRShare\% (standard deviation
\mOneRShSd{} points). Calibration selects $k=1$ in \mKcalOne{} of 100
splits. The features carry real signal: random assignment at the same
share reaches \mRand{} and an oracle that ranks series by their true
FM advantage reaches \mOrc, and the rule closes \mGap\% of that gap. Among
series it routes to the FM, the FM is better on \mPrec\% against \mRest\%
for those it does not route.

\paragraph{Higher votes are less reliable}
A 2-of-4 vote differs from always-FM by only $\mTwoD$ on the headline
split (95\% CI $[\mTwoLo, \mTwoHi]$, which includes zero) and
by $\mTwoRD$ on average over splits, and it beats always-FM in only
\mTwoFrac\% of them. Its FM share also swings from \mTwoRPlo\% to \mTwoRPhi\%
across splits (5th to 95th percentile), because the thresholds it
relies on are fit on few borderline series. A 3-of-4 vote is worse than
always-FM on average over splits.

\paragraph{What the four features add}
Seasonal autocorrelation alone differs from always-FM by $\mSACD$
(95\% CI $[\mSACLo, \mSACHi]$) with \mSACShare\% FM share, and in
\mSACFrac\% of splits, close to the 1-of-4 vote. Spectral entropy and CV
alone are unreliable: they beat always-FM in \mSEFrac\% and \mCVFrac\% of
splits. The vote therefore adds only a small, more reliable gain over
the best single feature, and we do not claim that all four are needed.

\paragraph{Comparison with a learned router}
At its default cutoff the logistic regression sends \mLRShare\% of series
to the FM and differs from always-FM by $\mLRD$ (95\% CI
$[\mLRLo, \mLRHi]$), or $\mLRRD$ on average over splits. The extra gain over the 1-of-4 rule
comes from spending more FM calls. At the same FM share as the 1-of-4
rule its difference is $\mLRmD$ (95\% CI $[\mLRmLo, \mLRmHi]$), no better
than the rule: averaged over splits, its MASE minus the rule's is
$\mLrDiff$ (standard deviation \mLrDiffSd) and it is lower in only
\mLrBetter\% of splits. On M4, therefore, the learned combiner does not beat the
simple rule at equal cost; it offers a different operating point.
Fig.~\ref{fig:pareto} plots MASE against the share of series routed to
the FM for all methods: the 1-of-4 rule and the logistic cutoff sweep lie
close together, and both sit well above the oracle curve.

\subsection{Mixed Portfolio: Traffic and M4 Daily}
Table~\ref{tab:pooled} reports the mixed portfolio ($n=\pN$ holdout). On
Traffic the FM is much more accurate, so the portfolio's FM-favorable
share is \pFmWin\% and always-FM beats always-specialist by \pFmSpec{}
MASE. Here the logistic regression beats the vote rule at equal FM share. At its
default cutoff it sends \pLRShare\% of series to the FM and differs from
always-FM by $\pLRD$ (95\% CI $[\pLRLo, \pLRHi]$). At the same FM
share as the 1-of-4 rule, its MASE minus the rule's is $\pLrDiff$ on
average over splits (standard deviation \pLrDiffSd), and it is lower in
\pLrBetter\% of splits. The 1-of-4 rule sends
\pTrafficKOneSh\% of Traffic series and \pMFourKOneSh\% of M4 series to the FM.
On average over splits it differs from always-FM by $\pOneRD$ (better
in \pOneFrac\% of splits), but the headline split alone is not conclusive
($\pOneD$, 95\% CI $[\pOneLo, \pOneHi]$). Higher votes lose accuracy.
In this experiment, dataset identity is withheld from the router so that
both methods share a single global rule across sources (whereas dispatch
still selects each dataset's calibrated model). A plausible reading is
that on a mixed portfolio the dataset-level signal in the features
matters, and a classifier combines it better than a single shared vote
threshold.

\begin{table*}[t]
\caption{Mixed Traffic and M4 Daily portfolio: holdout of the headline
split ($n=\pN$) and 100 repeated splits. Columns as in
Table~\ref{tab:m4}. Single features are omitted.}
\label{tab:pooled}
\centering
\footnotesize
\setlength{\tabcolsep}{5pt}
\begin{tabular}{lrrlrr}
\toprule
\textbf{Method} & \textbf{FM share (\%)} & \textbf{MASE} & \textbf{$\Delta$ vs.\ always-FM [95\% CI]} & \textbf{$\Delta$, 100 splits} & \textbf{Better (\%)} \\
\midrule
Always specialist (DLinear) & 0.0 (0.0) & 1.205 & $+0.118$ [$+0.101$, $+0.136$] & $+0.111$ (0.007) & 0 \\
Always FM & 100.0 & 1.086 & --- & --- & --- \\
\midrule
Router, 1-of-4 vote & 50.3 (6.0) & 1.081 & $-0.005$ [$-0.015$, $+0.005$] & $-0.014$ (0.005) & 99 \\
Router, 2-of-4 vote & 27.5 (8.1) & 1.093 & $+0.006$ [$-0.008$, $+0.021$] & $-0.001$ (0.011) & 37 \\
Router, 3-of-4 vote & 20.3 (4.6) & 1.105 & $+0.019$ [$+0.004$, $+0.035$] & $+0.011$ (0.011) & 12 \\
Router, 4-of-4 vote & 12.6 (1.4) & 1.146 & $+0.060$ [$+0.044$, $+0.077$] & $+0.055$ (0.010) & 0 \\
\midrule
Logistic regression ($p\geq0.5$) & 52.0 (4.3) & 1.060 & $-0.026$ [$-0.036$, $-0.016$] & $-0.025$ (0.004) & 100 \\
Logistic regression (same FM share as 1-of-4) & 50.3 (6.0) & 1.059 & $-0.028$ [$-0.038$, $-0.018$] & $-0.024$ (0.004) & 100 \\
\bottomrule
\end{tabular}

\end{table*}

\subsection{Cost Model and Break-Even}
\label{sec:cost}
Expected compute cost per series, in units of one specialist call, is
\begin{equation}
  E[C] = \alpha \cdot C_{\text{FM}} + (1-\alpha) \cdot C_{\text{spec}}
         + \varepsilon,
\end{equation}
where $\alpha$ is the fraction of series routed to the FM and
$\varepsilon$ is the router's own feature-extraction cost. Feature
extraction takes about 0.9\,ms on an M4-length history and DLinear takes
0.13\,ms (Table~\ref{tab:throughput}), which gives
$\varepsilon \approx \costEps$ specialist calls. The two timings come from
different machines, so this is approximate. Routing is cheaper than always-FM when $C_{\text{FM}} / C_{\text{spec}} >
1 + \varepsilon/(1-\alpha)$; for the 1-of-4 rule on M4
($\alpha \approx \mOneRShareShort\%$) the break-even ratio is about
\costBE{} at $\varepsilon=\costEps$. Table~\ref{tab:cost} gives expected cost as a fraction of
always-FM. At a ratio of 10 the router with overhead costs 1.07 of always-FM, slightly
more, at 100 it costs 0.39 of it and at 1{,}000 it costs 0.32. The measured
throughput gap is far larger than this break-even ratio, so on the hardware
pairs we measured the router saves cost; it would not with a cheap FM or
well-batched GPU inference at a ratio near 10. The
figures leave out two costs that amortize with volume: running the FM on
calibration series once, and training the specialist.

\begin{table}[t]
\caption{Expected cost as a fraction of always-FM cost on M4 Daily, at
the mean FM share over 100 splits. Overhead is the router's
feature-extraction cost, taken as $\varepsilon = 0.9/0.13 \approx 6.9$ ($\approx \costEps$) specialist calls;
``w/o'' excludes it and ``with'' includes it.}
\label{tab:cost}
\centering
\footnotesize
\begin{tabular}{lrrrr}
\toprule
 & \multicolumn{2}{c}{\textbf{Router, 1-of-4}} & \multicolumn{2}{c}{\textbf{Log.\ reg.\ ($p\geq0.5$)}} \\
\textbf{$C_{\text{FM}}/C_{\text{spec}}$} & \textbf{w/o} & \textbf{with} & \textbf{w/o} & \textbf{with} \\
\midrule
10$\times$ & 0.38 & 1.07 & 0.56 & 1.25 \\
100$\times$ & 0.32 & 0.39 & 0.51 & 0.58 \\
1{,}000$\times$ & 0.31 & 0.32 & 0.51 & 0.51 \\
\bottomrule
\end{tabular}

\end{table}

\subsection{Oracle Upper Bound}
To bound the headroom we compute an oracle that is not deployable: it
ranks holdout series by their true per-series FM advantage, which
requires the realized errors of both classes, and sends the top
fraction to the FM. On M4 its best point is at an FM share of \mOrcBestA{}
with MASE \mOrcBest, against \mOrc{} at the 1-of-4 rule's own share and
\mOneMase{} for the rule. A substantial gap remains, which reflects the
limit of what four coarse history features reveal about which individual
series the FM will forecast better.

\subsection{Implementation}
The router is a lightweight Python module that runs before model
dispatch. Evaluation code and the decile analysis are available at
\url{https://github.com/kavin-soni/timeseries-zeroshot-eval}.

\section{Limitations}
Within-dataset routing is evaluated on a single dataset, M4 Daily, with
a short 30-step FM context that may understate FM accuracy; Traffic is
FM-favorable on nearly every series, and ETTh1 and Exchange are too
small to route. We have not audited pretraining overlap between the FMs
and M4 or Traffic, nor evaluated newer checkpoints such as TimesFM~3.0.
We scope the candidate pool to 200M-parameter FMs and globally trained
neural and linear specialists (where PatchTST underperformed, leaving
DLinear); per-series statistical methods (ETS, Theta) and smaller
distilled FMs lie outside this scope and could shift both accuracy and
cost.

Methodologically, specialist training and Chronos sampling were run once
without fixed seeds; because run-to-run specialist variance on Traffic
($\approx 0.05$ MASE) exceeds the M4 FM-versus-DLinear gap (\mFmSpec),
retraining could shift the FM-favorable share. Each series is evaluated
at one forecast origin using mean MASE alone, splits are across series
rather than time, and overlapping splits plus independent bootstrapping
understate portfolio-level uncertainty. Finally, the router sees the
full training history rather than the $L$-step FM window, thresholds
must be recalibrated per portfolio and model pair, and throughput was
measured on separate hardware without GPU batching.

\section{Conclusion}
\label{sec:conclusion}
Routing each series to a foundation model or a cheap specialist before
inference, using four features of its history, can save FM calls without
costing accuracy where the two are close. On M4 Daily a 1-of-4 vote rule
improves on always-FM in all 100 repeated routing splits while sending
about a third of series to the FM, and a logistic regression on the same
features matches it at equal cost. On a mixed Traffic and M4 portfolio
the learned router beats the vote rule at equal cost, though part of
that gain may come from separating the two datasets. Savings depend on
the cost ratio and appear once the FM costs more than roughly eleven
specialist calls when the router's overhead is counted. Beyond
expanding the empirical scope noted above, future work includes
multi-tier routing across more than two model classes and value-weighted
uplift objectives or interpretable tree-based
routers~\cite{guduguntla2026valor} when accuracy on high-value series
matters more than unweighted win rates.

\section*{Acknowledgment}
In line with IEEE policy, we disclose that the AI systems Claude
(Anthropic) and Gemini (Google) were used in preparing this paper. They assisted with the
analysis code that produced the tables and figures in
Sections~\ref{sec:protocol} and~\ref{sec:router} (including
Figs.~\ref{fig:regimemap}--\ref{fig:deciles}), with editing the grammar of text in all sections, and
with \LaTeX{} formatting. Their role was assistive: the authors conceived and
designed the study, made all methodological decisions, reviewed and
verified all AI-assisted code, results and text, and take full
responsibility for the content.

\bibliographystyle{IEEEtran}
\bibliography{references}

\end{document}